\documentclass{article}

\usepackage{times}
\usepackage{latexsym}
\usepackage{url}
\usepackage{amsmath}
\usepackage{tabularx}
\usepackage{booktabs}
\usepackage{amssymb}
\usepackage{pifont}
\usepackage{bm}
\usepackage{multirow}
\usepackage{tcolorbox}
\usepackage[framemethod=tikz]{mdframed}
\usepackage{tikz,pgfplots}
\usepgfplotslibrary{groupplots}
\usepackage{enumitem}
  {\begin{enumerate}≈ \footnotesize%
    \setlength{\itemsep}{0pt}%
    \setlength{\parskip}{0pt}%
    }%
  {\end{enumerate}}

\newenvironment{tightitemize}%
  {\begin{itemize}[topsep=0pt, partopsep=0pt] %
    \setlength{\itemsep}{0pt}%
    \setlength{\parskip}{0pt}%
    }%
  {\end{itemize}}

\usepackage{caption}

\usepackage{subcaption}
\usepackage{microtype}
\usepackage{graphicx}
\usepackage{booktabs} 
\usepackage{comment}
\usepackage{bm}
\usepackage{hyperref}
\usepackage{url}
\usepackage{parskip}
\usepackage{amsmath, amsfonts, amsthm}
\usepackage{tikz,pgfplots}

\definecolor{g-red}{HTML}{DB4437}
\definecolor{g-blue}{HTML}{4285F4}
\definecolor{g-green}{HTML}{0F9D58}
\definecolor{g-yellow}{HTML}{F4B400}
\definecolor{g-orange}{HTML}{FF9800}
\definecolor{g-grey}{HTML}{9E9E9E}
\definecolor{shannon}{HTML}{304FFE}
\definecolor{uw}{RGB}{138,43,226}
\definecolor{stanford}{RGB}{255,69,0}
\definecolor{const}{RGB}{68, 110, 182}
\definecolor{head}{RGB}{246, 180, 32}
\definecolor{freq}{RGB}{0, 0, 0}

\usepackage[accepted]{icml2020}
\begin{document}

\twocolumn[
\icmltitle{Description Based Text Classification with Reinforcement Learning}

\icmlsetsymbol{equal}{*}

\begin{icmlauthorlist}
\icmlauthor{Duo Chai}{shannon}
\icmlauthor{Wei Wu}{shannon}
\icmlauthor{Qinghong Han}{shannon,zju}
\icmlauthor{Wu Fei}{zju}
\icmlauthor{Jiwei Li}{shannon}
\vskip 0.3in
\end{icmlauthorlist}

\icmlaffiliation{shannon}{Shannon.AI}
\icmlaffiliation{zju}{Zhejiang University}

\icmlcorrespondingauthor{Jiwei Li}{jiwei\_li@shannonai.com}
]

\printAffiliationsAndNotice{}
\begin{abstract}
The task of text classification is usually divided into two stages: {\it text feature extraction} and {\it classification}. 
In this standard formalization,
categories are
merely represented as indexes in the label vocabulary, and 
the model
lacks for explicit instructions on what to classify. 
 Inspired by the current trend of formalizing NLP problems as question answering tasks, we propose a new framework for text classification, in which each
 category 
  label is associated with a category description.
 Descriptions are generated by hand-crafted templates or using abstractive/extractive models 
 from
  reinforcement learning. The concatenation of the description and the text is fed to the classifier to decide whether or not the current label should be assigned to the text. 
 The proposed strategy forces the model to 
attend to the most salient texts with respect to the label description, which can be regarded as a hard version of attention, leading to better performances. 
We  observe significant performance boosts over strong baselines on
a wide range of 
 text classification  tasks including single-label classification, multi-label classification and multi-aspect sentiment analysis. 
\end{abstract}

\section{Introduction}
Text classification \cite{kim-2014-convolutional,joulin2016bag,yang2016hierarchical} is a fundamental problem in
natural language processing. 
The task is to assign one or multiple category label(s) to a sequence of text tokens. 
 It has broad applications  such as sentiment analysis \cite{pang2002thumbs,maas2011learning,socher2013recursive,tang2014learning,tang2015document},
 aspect sentiment classification \cite{jo2011aspect,tang2015effective,wang2015rating,nguyen2015phrasernn,tang2016aspect,pontiki2016semeval,sun2019utilizing}, 
 topic classification \cite{schwartz1997maximum,quercia2012tweetlda,wang2012baselines}, 
 spam detection \cite{ott2011finding,ott2013negative,li2014towards}, etc.

Standardly, text classification is divided into the following two steps: (1)
{\it text feature extraction}: a sequence of texts is  mapped to a 
feature
representation based on handcrafted features such as bag of words \cite{pang2002thumbs}, topics \cite{blei2003latent,mcauliffe2008supervised}, or distributed vectors 
using neural models such as LSTMs \cite{hochreiter1997long}, CNNs \cite{kalchbrenner2014convolutional,kim-2014-convolutional} or recursive nets \cite{socher2013recursive,irsoy2014deep,li2015tree,bowman2016fast};
and (2) {\it classification}: the extracted representation is fed to a classifier such as SVM, logistic regression or the softmax function to output the category label. 

This standard formalization for the task of text classification has an  intrinsic drawback: categories are
merely represented as indexes in the label vocabulary, and
 lack for explicit instructions on what to classify.  Labels can only influence the training process when the supervision signals are back propagated to feature vectors extracted from  the feature extraction step.
Class indicators in the text, which might just be one or two keywords, could be deeply buried in the huge chunk of text, making it hard for the model to separate grain from chaff.
Additionally, signals for different classes might entangle in the text. 
Take the task of aspect sentiment classification \cite{lei2016rationalizing} as an example, the goal of which is to classify the sentiment of a specific aspect of a review. 
A review  might contain diverse sentiments towards different aspects and that they are entangled together, {\it e.g. ``clean updated room. friendly efficient staff . rate was too high.''}. Under the standard formalization,  
the label of a text sequence is merely an index indicating the sentiment of a predefined but not explicitly mentioned aspect from the view of the model.
The model needs to first learn to associate the relevant text with the target aspect, 
and then decide the sentiment, which inevitably adds to the difficulty. 

Inspired by the current trend of formalizing
NLP problems as question answering tasks \cite{levy-etal-2017-zero,mccann2018natural,li2019unified,li-etal-2019-entity,gardner2019question,raffel2019exploring}, 
we propose a new framework for text classification by formalizing it as a SQuAD-style machine reading comprehension task.
The key point for this formalization is to associate each class with a class description to explicitly tell the model what to classify. 
For example, the task of classifying hotel reviews with positive location in
aspect sentiment classification for review $\bm{x} =\{x_1,x_2,...,x_n\}$ is transformed to assigning a ``yes/no''  label to ``{[CLS] positive location [SEP]} $\bm{x}$", indicating whether
the attribute
towards the location of the hotel
 in 
 review 
$\bm{x}$ is positive. 
By explicitly mentioning what to classify, 
the incorporation of class description
forces the model to 
attend to the most salient texts with respect to the label, which can be regarded as a hard version of attention.
This strategy
provides a straightforward resolution to the  issues mentioned in the previous paragraph.

One key issue with this method is how to obtain category descriptions. 
Recent models that cast NLP problems as QA tasks \cite{li2019unified,li-etal-2019-entity,gardner2019question}  use
hand-crafted
 templates to generate descriptions, and have two major drawbacks: (1) it is labor-intensive to 
 predefine descriptions for each category, especially when the number of category is large; and (2)
the model performance is sensitive to how the descriptions are constructed and human-generated templates might be sub-optimal. 
To handle this issue, we propose to automatically generate descriptions 
using reinforcement learning. 
The description can be generated in an extractive way, extracting a substring of the input text and using it as the description,
or in an abstractive way, using generative model to generate a string of tokens and  using it as the description. 
The model is trained in an end-to-end fashion to jointly learn to generate proper class descriptions and to assign correct class labels to texts. 

We are able to observe significant performance boosts against strong baselines on
a wide range of 
 text classification benchmarks including single-label classification, multi-label classification and multi-aspect sentiment analysis. 
 
 
\section{Related Work}
\label{related_work}
\subsection{Text Classification}
Neural models such as CNNs \cite{kim-2014-convolutional}, LSTMs \cite{hochreiter1997long,tang-etal-2016-effective}, recursive nets \cite{socher2013recursive} or Transformers \cite{NIPS2017_7181,devlin-etal-2019-bert},
have been shown to be effective in text classification. 
\citet{joulin-etal-2017-bag, TACL999} proposed fastText, representing the whole text using the average of embeddings of constituent words. 

There has been work investigating the rich information behind class labels. 
In the literature of zero-shot text classification, knowledge of labels are incorporated in the form of word embeddings \cite{yogatama2017generative,rios-kavuluru-2018-shot}, or class descriptions \cite{zhang-etal-2019-integrating,srivastava-etal-2018-zero}.
\citet{wang-etal-2018-joint-embedding} proposed a label-embedding attentive model that jointly embeds words and labels in the same latent space, and the text representations are constructed directly using the text-label compatibility. \citet{sun-etal-2019-utilizing} constructed auxiliary sentences from the aspect in the task of aspect based sentiment analysis (ABSA) by using four different sentence templates, and thus converted ABSA to a sentence-pair classification task. \citet{wang-etal-2019-aspect} proposed to frame ABSA towards question answering (QA), and designed an attention network to select aspect-specific words, which alleviates the effects of noisy words for a specific aspect. 
Descriptions in 
\citet{sun-etal-2019-utilizing}  and \citet{wang-etal-2019-aspect}  are generated from crowd-sourcing. This work takes a major step forward, in which the model is able to learn to automatically generate proper label descriptions from reinforcement learning.

\subsection{Formalizing NLP Tasks as Question Answering}
\paragraph{Question Answering}
MRC models  \cite{rajpurkar-etal-2016-squad,seo2016bidirectional,wang2016multi,wang2016machine,xiong2016dynamic,xiong2017dcn,wang2016multi,shen2017reasonet,chen2017reading,rajpurkar-etal-2018-know} extract answer spans from passages given questions. 
The task can be formalized as two multi-class classification tasks, i.e., predicting the starting and ending positions of the answer spans given questions. 
The context can either be prepared in advance 
 \cite{DBLP:conf/iclr/SeoKFH17} or selected from a large scale open-domain corpus such as Wikipedia \cite{chen-etal-2017-reading}. 
\paragraph{Query Generation}
In the standard version of MRC QA systems, queries are defined in advance. 
Some of recent works have studied how to generate queries 
 for better answer extraction. \citet{yuan-etal-2017-machine}  combines supervised learning and reinforcement learning to generate natural language descriptions; \citet{yang-etal-2017-semi} trained a generative model to generate queries based on unlabeled texts to train QA models; \citet{du-etal-2017-learning} framed the task of description generation as a {\it seq2seq} task, where descriptions are generated  conditioning on the texts; \citet{zhao-etal-2018-paragraph}  utilized the copy mechanism \cite{gu-etal-2016-incorporating, NIPS2015_5866} and \citet{kumar2018putting} proposed a generator-evaluator framework that directly optimizes objectives. Our work is similar to \citet{yuan-etal-2017-machine} and \citet{kumar2018putting} in terms of description generation, 
 in which reinforcement learning is applied for description/query generation. 
 
\paragraph{Formalizing NLP tasks as QA}
There has recently been a trend of casting NLP problem as QA tasks. 
\citet{gardner2019question} posed three motivations for using question answering as a format for a particular task, 
{\it i.e.}, to fill human information needs, to probe a system's understanding of some context and to transfer learned parameters from one task to another. , 
\citet{levy-etal-2017-zero} transformed the task of relation extraction to a QA task, in which each relation type $r(\bm{x},\bm{y})$ is characterized as a question ${q}(\bm{x})$ whose answer is $\bm{y}$. 
In a followup, \citet{li-etal-2019-entity} formalized the task of  entity-relation extraction as a multi-turn QA task by utilizing a template-based procedure to construct descriptions for relations and extract pairs of entities between which a relation holds. 
\citet{li2019unified} introduced a QA framework for the task of 
 named entity recognition, in which the extraction of an entity within the text is formalized as answering questions like  "{\it which person is mentioned in the text?}". 
\citet{mccann2018natural} built a multi-task question answering network for different NLP tasks, for example, the generation of a summary given a chunk of text is formalized as answering the question ``{\it What is the summary?}''. 
\citet{Wu2019CoreferenceRA} formalized the task of coreference as a question answering task.

\section{description-based Text Classification}
\label{query_based_text_classification}
Consider 
a sequence of text 
 $x=\{x_1,\cdots,x_L\}$ to classify, where $L$ denotes the length of the text $\bm{x}$. 
 Each $x$ is associated with a class label $y \in \mathcal{Y} = [1,N]$, where $N$ denotes the number of the predefined classes. 
 It is worth noting that in the task of single-label classification, $y$ can  take  only one value. While for the multi-label classification task, 
$y$ can take multiple values.

We use BERT \cite{devlin-etal-2019-bert}  as the backbone to illustrate how the proposed method works. 
It is worth noting that 
 the proposed method 
 is a general one and 
 can be easily extended to other model bases with minor adjustments. 
Under the formalization of 
the description-based text classification, each
class $y$ 
  is associated with a unique natural language description $\bm{q}_{y}=\{q_{y1},\cdots,q_{yL}\}$. 
The description encodes  prior information about the label and  facilitates the process of classification. 

For an N-class multi-class classification task, empirically, one can train  N binary classifiers or an N-class classifier, as will be separately described below.
\paragraph{N binary classifiers}
For the strategy of training N binary classifers, 
we iterate over all $q_y$ to decide whether the label $y$ should be assigned to a given instance $x$. 
More concretely, we first concatenate the text $\bm{x}$ and with the description $\bm{q}_y$ to generate $\{\text{[CLS]};\bm{q}_y;\text{[SEP]};\bm{x}\}$, where $\text{[CLS]}$ and $\text{[SEP]}$ are special tokens.
Next, the concatenated sequence
 is fed to transformers in BERT, from which we 
 we obtain the contextual representations $h_\text{[CLS]}$.
 Now that the representation $h_\text{[CLS]}$ has encoded interactions between the text and the description, another two-layer feed forward network is used to transform $h_\text{[CLS]}$ to a real value between 0 and 1 by using the sigmoid function, representing the probability of label $y$ being assigned to the text $\bm{x}$, as follows:
\begin{equation}
\label{eq:prediction}
p(y|\bm{x})=\text{sigmoid}(W_2\text{ReLU}(W_1h_\text{[CLS]}+b_1)+b_2)
\end{equation}
where $W_1,W_2,b_1,b_2$ are parameters to optimize. 
At test time,
for a multi-label classification task, in which multiple labels can be assigned to an instance, the resulting label set is as follows:
\begin{equation}
    \tilde{\bm{y}}=\{y~|~p(y|\bm{x})>0.5,\forall y\in\mathcal{Y}\}
\end{equation}
and for single-label classification, the resulting label set is as follows:
\begin{equation}
    \tilde{\bm{y}}=\arg\max_y(\{p(y|\bm{x}),\forall y\in\mathcal{Y}\})
\end{equation}

\paragraph{One N-class classifier}
For the strategy of training an N-class classifier, 
we concatenate all descriptions with the input $x$, which is given as follows:
$$\{\text{[CLS1]};\bm{q}_1;\text{[CLS2]};\bm{q}_2;...;\text{[CLS-N]};\bm{q}_N;\text{[SEP]};\bm{x}\}$$ 
where $\text{[CLSn]}~1\leq n\leq N$ are the special place-holding tokens. 
The concatenated input is then fed to the transformer, from which we obtain the 
 the contextual representations $h_\text{[CLS1]}, h_\text{[CLS2]}, ...,h_\text{[CLSN]}$. 
 The probability of assigning class $n$ to instance $x$ is obtained by first mapping $h_\text{[CLSn]}$ to scalars, and then outputting them to a softmax function,
   which
 is given as follows:
 \begin{equation}
 \begin{aligned}
& a_n = \hat{h}^T\cdot h_\text{[CLSn]} \\
& p(y=n|x) = \frac{\exp{(a_n)}}{\sum_{t=1}^{t=N} \exp{(a_t)}}
 \end{aligned}
 \end{equation}
It is worth noting that the on N-class classifier strategy can not handle the multi-label classification case. 

\section{Description Construction}
\label{query_construction}
\begin{figure*}[th]
    \centering
    \includegraphics[scale=0.6]{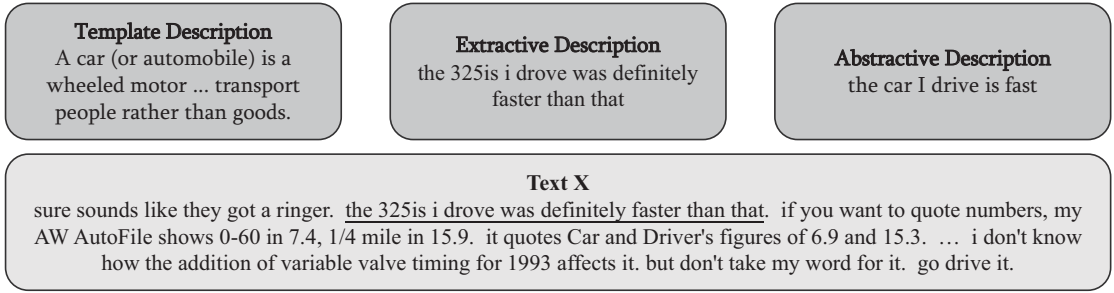}
    \vskip 0.1in
    \caption{An example  of descriptions constructed via different strategies. Text is from the {20news} dataset.}
    \label{fig:overview}
\end{figure*}

In this section, we describe the three
proposed
 strategies to construct descriptions:  the  {\bf template (Tem)}  strategy (Section~\ref{temp}), the {\bf extractive  (Ext)} strategy (Section~\ref{ext}) and the {\bf abstractive  (Abs)} strategy (Section~\ref{abs}). An example of descriptions constructed by different strategies is shown in Figure~\ref{fig:overview}.

\subsection{The Template Strategy}
\label{temp}
As previous works \cite{ li-etal-2019-entity,li2019unified,levy-etal-2017-zero} did, the 
most straightforward 
 way to construct 
 label
 descriptions is to use handcrafted templates. 
 Templates can come from various sources, such as Wikipedia definitions, or human annotators. 
 Example explnanations
 for some of the 20 news categoroes
  are shown in Table~\ref{template}. More comprehensive template descriptions are listed in the supplementary material.
\begin{table*}
\label{tab:query}
\centering
\small
\vskip 0.1in
\begin{tabular}{ll}
    \toprule
    {\bf Label} & {\bf Description}\\
    \midrule
    {\sc comp.sys.mac.hardware} & {The Macintosh is a family of personal computers designed ... since January 1984.}\\
    {\sc rec.autos} & {A car (or automobile) is a wheeled motor ... transport people rather than goods.}\\
    {\sc talk.politics.misc} & {Politics is a set of activities ... making decisions that apply to groups of members.}\\
    \bottomrule
\end{tabular}
\caption{Examples of template descriptions drawn from Wikipedia for the \texttt{20news} dataset. For other labels and datasets, we also use their Wikipedia definitions as template descriptions.}
\label{template}
\end{table*}

\subsection{The Extractive Strategy}
\label{ext}
Generating descriptions using templates is suboptimal since (1) it is labor-intensive to ask humans to write down templates for different classes, especially when the number of classes is large; and (2) inappropriately constructed templates will actually lead to inferior performances, as demonstrated in \citet{li2019unified}.  The model should have the ability to learn to {\it generate} the most appropriate  descriptions regarding different classes 
conditioning on
the current text to classify, and the appropriateness of the generated descriptions should directly correlate with the final classification performance. 
To this end, we describe two ways to generate descriptions, the {\it extractive} strategy, as will be detailed in this subsection, and the {\it abstractive} strategy, which will be detailed in the next subsection.

For the extractive strategy, for each input $\bm{x}=\{x_1,\cdots,x_T\}$, 
the extractive model generates a description 
$\bm{q}_{yx}$ for each class label $y$, where $\bm{q}_{yx}$ is a substring of $\bm{x}$. 
As can be seen, for different inputs $x$, the descriptions for the same class can be different.
For the golden class  $y$ 
that should be assigned 
 to  $x$, 
there should be a substring of $x$  relevant to $y$, and this substring will be chosen as the description for $y$.
But  classes that should not be assigned, there might not be corresponding substrings in $x$ that can be used as descriptions. 
To deal with this issue, we append $N$ dummy tokens to $x$, providing the model the flexibility of handling the case where there is no corresponding substring within $x$ to a class label. 
If the extractive model picks a dummy token,
it will use  hand-crafted templates for different categories as descriptions. 
 
To back-propagate the signal indicating which span contributes how much to the classification performance, 
we turn to reinforcement learning, an approach that encourages the model to act toward higher rewards. A typical reinforcement learning algorithm consists of three components: the action $a$, the policy $\pi$ and the reward $R$.

\paragraph{Action and Policy} 
For each class label $y$, the action is to pick a text span $\{x_{i_s},\cdots,x_{i_e}\}$ from $\bm{x}$ to represent $\bm{q}_{yx}$. Since a span is a sequence of continuous tokens in the text, we only need to select the starting index $i_s$ and the ending index $i_e$, denoted by $a_{i_s,i_e}$.

For each class label $y$,
the policy $\pi$ defines the probability of selecting the starting index $i_s$ and the ending index $i_e$. 
Following previous work \cite{chen-etal-2017-reading,devlin-etal-2019-bert}, each token $x_k$ within $\bm{x}$ is mapped to a representation $h_k$ using BERT, and
the probability  of $x_i$ being the starting index and the ending index
of $\bm{q}_{yx}$
 are given as follows: 
\begin{equation}
\begin{aligned}
&    P_\text{start}(y, k) = \frac{\exp(W^{ys} h_k)}{\sum_{t=1}^{t=T} \exp(W^{ys} h_t) } \\
&    P_\text{end}(y, k) = \frac{\exp(W^{ye} h_k)}{\sum_{t=1}^{t=T} \exp(W^{ye} h_t) } \\
\end{aligned}
\end{equation}
where $W^{ys}$ and $W^{ye}$ are $1\times K$ dimensional vectors to map $h_t$ to a scalar. 
Each class $y$ has a class-specific  $W^{ys}$ and $W^{ye}$.
The probability of a text span with the starting index  $i_s$ and ending index $i_e$ being the description for class $y$ , denoted by $P_\text{span}(y, a_{i_s,i_e})$, is given as follows: 
\begin{equation}
    P_\text{span}(y, a_{i_s,i_e})=P_\text{start}(y, i_s)\times P_\text{end}(y, i_e)
\end{equation}


\paragraph{Reward}
Given $\bm{x}$ and a description $\bm{q}_{yx}$, the classification model in Section \ref{query_based_text_classification} 
will output the probability of assigning the correct label to $x$, which will be used as the reward to update both the 
classification model and the extractive model. 
Specifically, for multi-class classification, all $\bm{q}_{yx}$ are concatenated with $\bm{x}$, and the reward is given as  follows
\begin{equation}
    \label{eq:reward}
    R(x, \bm{q}_{yx} \text{for~all~y})= p(y=n|x)
\end{equation}
where $n$ is the gold label for $x$. 

For N-binary-classification model, each $\bm{q}_{yx}$  is separately concatenated with $x$, and the reward is given as follows:
\begin{equation}
R(x, \bm{q}_{yx})= p(y=\hat{y}|x)
\end{equation}
where $\hat{y}$ is the golden binary label. \footnote{Experiments show that using the probability as the reward performs better than using the log probability.}
\paragraph{REINFORCE}
To find the optimal policy, we use the REINFORCE algorithm \cite{DBLP:journals/ml/Williams92}, a kind of policy gradient method which maximizes the expected reward $\mathbb{E}_\pi[R(x, \bm{q}_y)]$. For each generated description $\bm{q}_{yx}$ and the corresponding $x$, we define its loss as follows:
\begin{equation}
    \label{eq:reinforce}
    \mathcal{L}=-\mathbb{E}_\pi[R(\bm{q}_{yx}, x)]
\end{equation}
REINFORCE approximates the expectation in Eq.~\ref{eq:reinforce} with sampled descriptions from the policy distribution. The gradient to update parameters is given as follows:
\begin{equation}
    \label{eq:gradient}
    \nabla\mathcal{L}\approx -\sum_{i=1}^B\nabla\log \pi(a_{i_s,i_e}|\bm{x},y)[R(\bm{q}_y)-b]
\end{equation}
where $b$ denotes the baseline value, which is set to the average of all previous rewards.
The extractive policy is initialized to generate dummy tokens as descriptions.
Then the extractive model and the classification model are jointly trained based on the reward.

\subsection{The Abstractive Strategy}
\label{abs}
An alternative generation strategy is to generate descriptions using generation models. 
The generation model uses the sequence-to-sequence structure \cite{sutskever2014sequence,NIPS2017_7181} as a backbone.
It takes $x$ as an input, and generate different descriptions $q_{yx}$ for different $x$.

\paragraph{Action and Policy}
For each class label $y$, the action
 is to generate the description $\bm{q}_{yx}=\{q_1,\cdots,q_L\}$, defined by $p_\theta$. 
The policy $P_\textsc{seq2seq}$ defines the probability of generating the entire string of the description given $x$, which 
 is equivalent to generating each token within the description, and is given as follows:
\begin{equation}
    P_\textsc{seq2seq}(\bm{q}_y|x)=\prod_{i=1}^L p_\theta(q_i|q_{<i},x,y)
\end{equation}
where $q_{<i}$ denotes all the already generated tokens. 
 $P_\textsc{seq2seq}(\bm{q}_y|x)$ 
for different class $y$ share the structures and parameters, with the only difference being that a class-specific embedding $h_y$ is appended to each source and target token.

\paragraph{Reward}
The RL reward  and the training loss for the abstractive  strategy are similar to those for the extractive strategy, as in  Eq.~\ref{eq:reward} and
 in Eq.~\ref{eq:reinforce}. 
  A widely recognized challenge for training language models using RL is the high variance, since the action space is huge \cite{ranzato2015sequence,yu2017seqgan,li-etal-2017-adversarial}.  
To deal with this issue, we use the REGS -- Reward for Every Generation Step proposed by \citet{li-etal-2017-adversarial}. 
Unlike standard REINFORCE training, in which the same reward is used to update the probability of all tokens within the description,
REGS trains a  a discriminator that
is able to assign rewards to partially decoded sequences.  
 The gradient is  given by:
\begin{equation}
    \label{eq:gradient2}
    \nabla\mathcal{L}\approx -\sum_{i=1}^L\nabla\log \pi(q_i|q_{<i},h_y)[R(q_{<i})-b(q_{<i})]
\end{equation}
Here $R(q_{<i})$ denotes the reward given the partially decoded sequence $q_{<i}$ as the description, and
$b(q_{<i})$ denotes the baseline. 

The generative policy $P_\textsc{seq2seq}$ is initialized using  a pretrained encoder-decoder with input being $x$ and output being 
template descriptions. 
Then the description generation model and the classification model are jointly trained based on the reward. 


\section{Experiments}
\label{experiments}

\begin{table*}[t]
\small
    \centering
    \caption{Test error rates on the AGNews, 20news, DBPedia, Yahoo, Yelp P and IMDB datasets for single-label classification. `--' means not reported results. `$\sharp$' means we take the best reported results from \citet{10.1007/978-3-030-32381-3_16}.}
    \vskip 0.1in
    \begin{tabular}{clcccccc}
    \toprule
    &{\bf Model}& AGNews  & 20news & DBPedia & Yahoo &YelpP & IMDB \\\midrule
     &Char-level CNN \cite{zhang2015character} & 8.5 &  -- &  1.4 & 28.8 & 4.4 & -- \\
     &VDCNN \cite{conneau2016very}             & 8.7 &  -- & 1.3  & 26.6 & 4.3 & -- \\
     & DPCNN \cite{johnson-zhang-2017-deep}& 6.9 & -- & 0.91 & 23.9 & 2.6 & --\\
     &Label Embedding  \cite{wang2018joint}    & 7.5 & --  & 1.0  & 22.6 & 4.7 & -- \\\midrule
     &LSTMs \cite{zhang2015character}          & 13.9 & 22.5 & 1.4 & 29.2  & 5.3  & 9.6 \\ 
     &Hierarchical Attention \cite{yang2016hierarchical} & 11.8 & 19.6 & 1.6 & 24.2 &5.0  & 8.0\\
     &D-LSTM\cite{yogatama2017generative} & 7.9 & -- & 1.3 & 26.3 & 7.4 & --\\
     &Skim-LSTM \cite{seo2018neural}& 6.4 & -- & -- & -- & -- & 8.8\\\midrule
     &BERT \cite{devlin2018bert} & \underline{5.9}  &\underline{16.9}&\underline{0.72}& \underline{22.7}& \underline{2.4} & \underline{6.8}\\
    \midrule
    &Description (Tem.) & 5.2 & 15.8 & 0.65 & 22.1& 2.2 & 5.8 \\
    &Description (Ext.) & {\bf 5.0} & 15.6 & {\bf 0.63} & 22.0&2.1 & {\bf 5.5}  \\
    &Description (Abs.) & 5.1 & {\bf 15.4} & 0.62 & {\bf 21.8}& {\bf 2.0} & {\bf 5.5}  \\\bottomrule
    \end{tabular}
    \label{res1}
\end{table*}

\subsection{Benchmarks}
We use the following widely used benchmarks to test the proposed model. The detailed descriptions for benchmarks are found in the supplementary material. 
\begin{tightitemize}
\item{\bf Single-label Classification}: The task of single-label classification is to assign a single class label to the text to classify. We use the following widely used benchmarks: (1) {\bf AGNews}: Topic classification over four categories of Internet news articles \cite{del2005ranking}.
(2)  {\bf 20newsgroups}\footnote{\url{http://qwone.com/~jason/20Newsgroups/}}: The 20 Newsgroups data set is a collection of  documents over 20 different newsgroups. 
(3) {\bf DBPedia}: Ontology classification over fourteen non-overlapping classes picked from
DBpedia 2014 (Wikipedia).
(4) {\bf Yahoo! Answers}: Topic classification  ten largest main categories from Yahoo! Answers. 
(5) {\bf Yelp Review Polarity (YelpP)}: 
Binary sentiment classification over  yelp reviews. 
(6) {\bf IMDB}: Binary sentiment classification over IMDB reviews. 
\item{\bf Multi-label Classification}: The goal of multi-label classification is to assign multiple class labels to a single text. We use (1) {\bf Reuters}\footnote{\url{https://martin-thoma.com/nlp-reuters/}}: A multi-label benchmark dataset for document classification with 90 classes.   
 (2) {\bf AAPD}: The arXiv Academic Paper dataset \cite{yang2018sgm} with  54 classes. 
\item {\bf Multi-aspect Sentiment Analysis}: 
The goal of the task is to test a model's ability to 
identify entangled sentiments for different aspects 
of a review.
Each review might contain diverse sentiments towards different aspects. 
Widely used datasets include (1) the {\bf BeerAdvocate} review dataset  over  aspects  \texttt{appearance}, \texttt{smell}.
  \citet{lei2016rationalizing} processed the dataset by picking examples with  less correlated aspects, leading to 
a de-correlated subset for each aspect
(aroma) and \texttt{palate}. 
(2) the hotel {\bf TripAdvisor} review \cite{li2016understanding} over four aspects, {\it i.e.},   \texttt{service}, \texttt{cleanliness}, \texttt{location} and \texttt{rooms}. \citet{li2016understanding} processed the dataset by  picking examples with  less correlated aspects.
There are three classes, positive, negative and neutral for both datasets.
\end{tightitemize}


    
\begin{table}[t]
    \centering
    \small
    \caption{Test error rates on the Reuters and AAPD datasets for multi-label classification.}
    \vskip 0.1in
    \begin{tabular}{ccc}
    \toprule
    {\bf Model}& Reuters & AAPD \\\midrule
    LSTMs \cite{zhang2015character} & 16.8&33.5 \\ 
    Hi-Attention \cite{yang2016hierarchical} & 13.9 & 30.3 \\
    Label-Emb \cite{wang2018joint} &13.6  & 29.9 \\
    LSTM$_{reg}$ \cite{adhikari-etal-2019-rethinking} & 13.0 & 29.5\\
    BERT \cite{adhikari2019docbert}  & \underline{11.0}  & \underline{26.6} \\\midrule
    Description (Tem.) & 10.3 & 25.9 \\
        Description (Ext.)  & 10.1 & 26.0 \\
    Description (Abs.) & {\bf 10.0} & {\bf 25.7} \\\bottomrule
    \end{tabular}
    \label{res-multi}
\end{table} 
    
\begin{table}[t]
    \centering
    \small
    \caption{Test error rates on the BeerAdvocate (Beer), TripAdvisor (Trip) for multi-aspect sentiment classification.}
    \vskip 0.1in
    \begin{tabular}{cccc}
    \toprule
    {\bf Model}& Beer & Trip  \\\midrule
    LSTMs \cite{zhang2015character} &   34.9 & 47.6 \\ 
    Hi-Attention \cite{yang2016hierarchical} & 33.3 & 42.2 \\
    Label-Emb  \cite{wang2018joint} &32.0  & 43.5 \\
    BERT \cite{devlin2018bert}  & \underline{27.8}  & \underline{35.6} \\\midrule
    Description (Tem.) & 17.4 & 18.1 \\
    Description (Ext.) & 16.0 & {\bf 17.0} \\
    Description (Abs.)  & {\bf 15.6} & 17.6 \\\bottomrule
    \end{tabular}
    \label{res-aspect}
\end{table}

 \subsection{Baselines}
 We implement the following widely-used models as baselines.
 Hyper-parameters for baselines are tuned on the development sets to
  enforce apple-to-apple comparison.
 In addition, we also copy results of models from  relevant papers. 
  \begin{itemize}
  \item {\bf LSTM}: The vanilla LSTM model \cite{zhang2015character}, which first maps the text sequence to a vector using LSTMs \cite{hochreiter1997long}. 
  For single-label datasets, the obtained document embeddings are output to the softmax layer. 
      For multi-label datasets, we follow \citet{adhikari2019docbert}, in which
  each label is associated with a binary sigmoid function, and then  
  the document embedding is fed to output the class label.   
  \item {\bf Hierarchical Attention} \cite{yang2016hierarchical}: The hierarchical attention model which uses word-level attention to obtain sentence embeddings and uses sentence-level attention to obtain document embeddings. 
  We follow the strategy adopted in the LSTM model to handle multi-label tasks. 
  \item {\bf Label Embedding} : Model proposed by \citet{wang2018joint} that jointly learns the label embeddings and document embeddings.  
  \item {\bf BERT}: We use the BERT-base model \cite{devlin2018bert,adhikari2019docbert} 
    as the  baseline. We follow the standard classification setup in BERT, in which the embedding of [CLS] is fed to a softmax layer to output the probability 
  of a class being assigned to an instance.  We follow the strategy adopted in the LSTM model to handle multi-label tasks. 
  \end{itemize}
  
\subsection{Results and Discussion}
Table~\ref{res1} presents the results  for   single-label classification tasks.
The three proposed strategies consistently outperform the BERT baseline. 
Specifically, the template-based strategy 
outperforms BERT, decreasing error rates by 
 {\it i.e.,} -0.7 on AGNews, -1.1 on 20news, -0.07 on DBPedia, -0.6 on Yahoo, -0.2 on YelpP and -1.0 on IMDB. 
The extractive and abstractive strategies consistently outperform the template-based strategy, which is because of their ability to automatically learn the proper descriptions. 
The extractive strategy performs better than the abstractive strategy on the AGNews and IMDB, but  worse on the others.

Table~\ref{res-multi} shows the results on the two multi-label classification datasets -- Reuters and AAPD. Again, we observe  performance gains over the BERT baseline on both datasets in terms of F1 score.

Table~\ref{res-aspect} shows the experimental results on the two multi-aspect sentiment analysis datasets BeerAdvocate and TripAdvisor. Surprisingly huge gains are observed on both datasets. Specifically, for BeerAdvocate, our method (Abs.) decreases the error rate from 27.8 to 15.6, and for TripAdvisor, our method (Ext.)
decreases the error rate from 35.6 to 17.0. 
The explanation for this huge  boost is as follows: 
 both datasets are deliberately constructed in a way that 
 each review contains 
   aspects with opposite sentiments entangling with each other. 
This makes it extremely hard for the model to learn to jointly identify  the  target aspect and the sentiment. 
The incorporation of description gives the model the ability to directly attend to the relevant text, which leads to significant performance boost. 
  
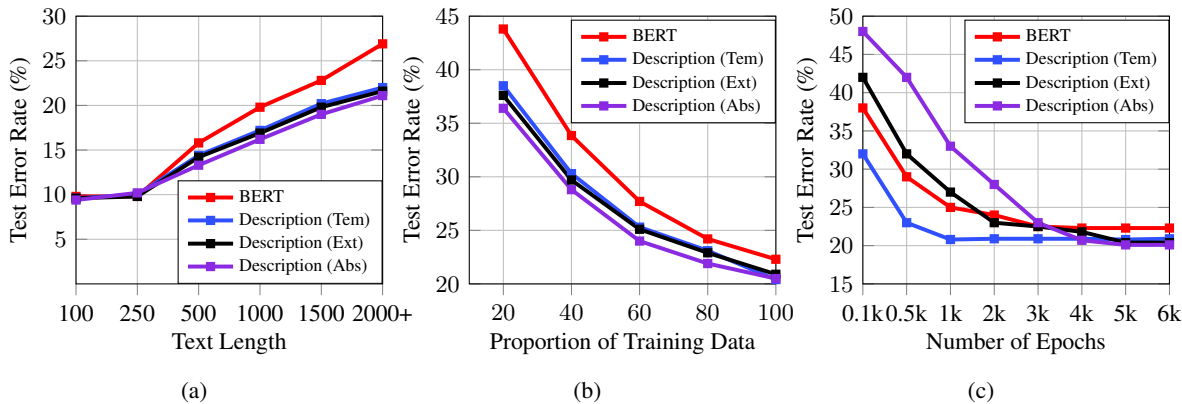
\begin{figure*}[!ht]     
     \begin{subfigure}{0.3\textwidth}
     \centering
\begin{tikzpicture}
\begin{axis}[
    	   width=1.1\columnwidth,
	    height=1\columnwidth,
	    legend cell align=left,
	    legend style={at={(1, 0)},anchor=south east,font=\small,nodes={scale=0.75, transform shape}},
	    xtick={1,2,3,4,5,6},
	    xticklabels={100, 250, 500, 1000, 1500,2000+},
   	 	ytick={5,10,15,20,25,30},
   		ymin=0, ymax=30,
   		xtick pos=left,
   		xtick align=outside,
	    xmin=1,xmax=6,
	    mark options={mark size=1},
		font=\small,
   	 	ymajorgrids=true,
    	xmajorgrids=true,
    	xlabel=Text Length,
        ylabel=Test Error Rate  (\%),
    	ylabel style={at={(axis description cs: 0.12, 0.5)}}]
    	
\addplot[
    color=red,
    mark=square*,
    line width=1.5pt
    ]
    coordinates {
(1, 9.8)(2, 9.8)(3,15.8)(4,19.8)(5,22.8)(6,26.9)
    };
    \addlegendentry{BERT}

\addplot[
    color=shannon,
    mark=square*,
    line width=1.5pt
    ]
    coordinates {
(1, 9.6)(2, 10.0)(3,14.4)(4,17.2)(5,20.2)(6,22.0)
    };
    \addlegendentry{Description (Tem)}

\addplot[
    color=black,
    mark=square*,
    line width=1.5pt
    ]
    coordinates {
(1, 9.6)(2, 9.8)(3,14.2)(4,16.9)(5,19.8)(6,21.6)
    };
    \addlegendentry{Description (Ext)}

\addplot[
    color=uw,
    mark=square*,
    line width=1.5pt
    ]
    coordinates {
(1,9.4)(2, 10.2)(3,13.3)(4,16.2)(5,19.0)(6,21.1)
    };
    \addlegendentry{Description (Abs)}
\end{axis}
\end{tikzpicture}
         \caption{}
        \label{text-length}
     \end{subfigure}
          \begin{subfigure}{0.3\textwidth}
     \centering
\begin{tikzpicture}
\begin{axis}[
    	   width=1.1\columnwidth,
	    height=1\columnwidth,
	    legend cell align=left,
	    legend style={at={(1, 0.6)},anchor=south east,font=\small,nodes={scale=0.75, transform shape}},
	    xtick={0,20, 40, 60, 80, 100},
   	 	ytick={5,10,15,20,25,30,35,40,45,50,55},
   		ymin=20, ymax=45,
   		xtick pos=left,
   		xtick align=outside,
	    xmin=10,xmax=100,
	    mark options={mark size=1},
		font=\small,
   	 	ymajorgrids=true,
    	xmajorgrids=true,
    	xlabel=Proportion of Training Data,
        ylabel=Test Error Rate  (\%),
    	ylabel style={at={(axis description cs: 0.12, 0.5)}}]
    	
\addplot[
    color=red,
    mark=square*,
    line width=1.5pt
    ]
    coordinates {
(20, 43.8)(40, 33.85)(60,27.7)(80,24.2)(100, 22.3)
    };
    \addlegendentry{BERT}

\addplot[
    color=shannon,
    mark=square*,
    line width=1.5pt
    ]
    coordinates {
(20,38.5)(40, 30.3)(60,25.3)(80,23.1)(100, 20.4)
    };
    \addlegendentry{Description (Tem)}

   \addplot[
    color=black,
    mark=square*,
    line width=1.5pt
    ]
    coordinates {
(20,37.6)(40,29.7)(60,25.1)(80,22.9)(100, 20.9)
    };
    \addlegendentry{Description (Ext)}
 
\addplot[
    color=uw,
    mark=square*,
    line width=1.5pt
    ]
    coordinates {
(20, 36.4)(40, 28.8)(60,24.0)(80,21.9)(100,20.5)
    };
    \addlegendentry{Description (Abs)}
\end{axis}
\end{tikzpicture}
         \caption{}
         \label{proportion}
     \end{subfigure}
               \begin{subfigure}{0.3\textwidth}
     \centering
\begin{tikzpicture}
\begin{axis}[
    	   width=1.1\columnwidth,
	    height=1\columnwidth,
	    legend cell align=left,
	    legend style={at={(1, 0.6)},anchor=south east,font=\small,nodes={scale=0.75, transform shape}},
	    xtick={1,2,3,4,5,6,7,8},
	    xticklabels={0.1k, 0.5k, 1k, 2k, 3k, 4k, 5k, 6k},
   	 	ytick={15,20,25,30,35,40,45,50},
   		ymin=15, ymax=50,
   		xtick pos=left,
   		xtick align=outside,
	    xmin=1,xmax=8,
	    mark options={mark size=1},
		font=\small,
   	 	ymajorgrids=true,
    	xmajorgrids=true,
    	xlabel=Number of Epochs,
        ylabel=Test Error Rate  (\%),
    	ylabel style={at={(axis description cs: 0.12, 0.5)}}]
    	
\addplot[
    color=red,
    mark=square*,
    line width=1.5pt
    ]
    coordinates {
(1, 38)(2, 29)(3, 25)(4,24)(5,22.5)(6,22.3)(7,22.3)(8,22.3)
    };
    \addlegendentry{BERT}

\addplot[
    color=shannon,
    mark=square*,
    line width=1.5pt
    ]
    coordinates {
(1, 32)(2, 23)(3, 20.8)(4,20.9)(5,20.9)(6,20.9)(7,20.8)(8,20.9)
    };
    \addlegendentry{Description (Tem)}

\addplot[
    color=black,
    mark=square*,
    line width=1.5pt
    ]
    coordinates {
(1, 42)(2, 32)(3, 27)(4,23)(5,22.5)(6,21.8)(7,20.4)(8,20.4)
    };
    \addlegendentry{Description (Ext)}

\addplot[
    color=uw,
    mark=square*,
    line width=1.5pt
    ]
    coordinates {
(1, 48)(2, 42)(3, 33)(4,28)(5,23)(6,20.7)(7,20.1)(8,20.1)
    };
    \addlegendentry{Description (Abs)}
\end{axis}
\end{tikzpicture}
\caption{}
\label{convergence}
     \end{subfigure}
    \caption{(a) Test error rate vs text length (b) Test error rate vs proportion of training data (c) Test error rate vs the number of epochs. }
\end{figure*}

\section{Ablation Studies and Analysis}
In this section, we perform comprehensive ablation studies for better understand the model's behaviors. 
More examples of human-crafted descriptions and descriptions learned from RL are shown in the supplementary material. 
\subsection{Impact of Human Generated Templates}
How to construct queries has a significant influence
on the final results. In this subsection, 
we use the Yahoo! Answer dataset for illustration. 
We use
different ways to construct template descriptions and
test their influences. 
\begin{tightitemize}
\item {\bf Label Index}: the description  is the index of a class, {\it i.e.} ``one'', ``two'', ``three''. 
\item {\bf Keyword}: the description is the keyword extension of each category. 
\item {\bf Keyword Expansion}: we use Wordnet to retrieve the synonyms of keywords and the description is their concatenation. 
\item {\bf Wikipedia}: definitions drawn from Wikipedia. 
\end{tightitemize}
\begin{table}[!ht]
\small
\center
\caption{Results on 20news using different templates as descriptions.}
\vskip 0.1in
\begin{tabular}{ll}
\toprule 
{\bf Model} & {\bf Error Rate}\\ \midrule
BERT &  16.9 \\
Template Description (Label Index) & 16.8 (-0.1)\\
Template Description (Keyword) &  16.4 (-0.5)\\
Template Description (Key Expansion) &  16.2 (-0.7)\\
Template Description (Wiki)&   15.8 (-1.1)\\\bottomrule
\end{tabular}
\label{diff-template}
\end{table}

Results are shown in Table \ref{diff-template}. As can be seen, the performance is sensitive to the way that descriptions are constructed. 
The performance for label index is very close to that of the BERT baseline. This is because label indexes do not carry any semantic knowledge about classes.
One can think of the representations for label indexes similar to the vectors for different classes in the softmax layer, 
making the two models theoretically the same. 
Wikipedia outperforms Keyword since  descriptions from Wikipedia carry more  comprehensive semantic information for each class.

\subsection{Impact on Examples with Different Lengths}
It is interesting to see how differently the description-based models affect examples with different lengths. 
We use the IMDB dataset for illustrations since the IMDB dataset contains texts with more varying lengths. 
Since 
the model trained on the 
 full set already has super low error rate
  (around 4-5$\%$), we worry about the noise in comparison.  
We thus train different models on 20$\%$ of the training set, and test them on the  test sets split into different buckets by text length. 

Results are shown in Figure \ref{text-length}. As can be seen, the superiority of 
description-based models over vanilla ones is more obvious on long texts. This is in line with our expectation: we can treat the descriptions as a hard version of attentions, forcing the model to look at the most relevant parts. 
For longer texts, where grain is mixed with larger amount of chaff, this mechanism will immediately introduce performance boosts.
But for short texts, which is relatively easy for classification, both models can easily detect the relevant part and correctly classify it, making the gap smaller.

\begin{table}
\center
\small
\begin{tabular}{ccc}\hline
& Yahoo Answer & AAPD \\\hline
Template &   22.4 & 25.9 \\\hline
Ext  (dummy Init) & 22.2&26.0     \\
Ext  (ROUGE-L Init) & 25.3 &27.2 \\
Ext (random Init)  & 28.0  & 30.1  \\\hline
Abs (template Init)  & 22.0  & 25.7  \\ 
Abs (random Init)  & 87.9 & 78.4  \\ \hline
\end{tabular}
\caption{Error rates for different RL initialization strategies.}
\label{RL}
\end{table}

\subsection{Convergence Speed}
Figure \ref{convergence} shows the convergence speed of different models on the Yahoo Answer dataset.
For the description-based methods, the template model converges faster than the BERT baseline. This is because templates encode prior knowledge about  categories.
Instead of having the model learn to attend to the relevant texts, template-based methods force the model to pay attention to the relevant part.
Both the abstractive strategy and the extractive strategy 
converge slower than  the  template-based method and the BERT baseline. This is because it has to learn to generate the relevant description using reinforcement 
learning. Since the REINFORCE method is known for large variance, the model is slow to converge. 
The extractive strategy converges faster  than the abstractive strategy due to the smaller search space. 

\subsection{Impact of the Size of Training Data}
Since the description encodes prior semantic knowledge about categories, we expect that description-based methods work better with less training data. 
We trained different models on different proportions of the Yahoo Answer dataset, and test them on the original test set. 
From  Figure \ref{proportion}, we can see that
the gap between the BERT baseline and the description-based models is significantly larger with 20\% of training data and the gap is gradually narrowed down with increasing amount of training data. 
\subsection{Impact of RL Initialization Strategies}
We explore the effect of different initialization strategies for RL on the Yahoo! Answer and  AAPD datasets.
For the extractive strategy, we explore random initialization
and
the ROUGE-L strategy.
For the ROUGE-L strategy, 
 the description for the correct label is the  span that achieves the highest ROUGE-L score with respect to the template.  The ROUGE-L strategy is widely used 
for sudo-golden answer/summary extraction in training extractive models, 
when  golden answers are not substrings of the text in question answering \cite{nguyen2016ms} or golden summaries are not substrings of the input document \cite{kovcisky2018narrativeqa} for the task of summarization. The descriptions for incorrect labels are dummy tokens for  the ROUGE-L strategy.

Results are shown in Table \ref{RL}. As can be seen, generally, initialization  matters.
The extractive model is more immune to initialization strategies, even random initialization achieves acceptable performances. This is because of the smaller search space for extractive models relative to abstractive models. 
For the random initialization of the abstractive model, 
we are not
able to make it converge within a reasonable amount of time.

\section{Conclusion}
\label{conclusion}
We present a  description-based text classification method that generates class-specific descriptions 
to give  the model
 an explicit guidance of what to classify, which mitigates the issue of ``meaningless labels''. We develop three strategies to construct descriptions, {\it i.e.,}  the template-based strategy, the extractive strategy and the abstractive strategy. 
The proposed framework achieves significant performance boost on a wide range of classification benchmarks.

\bibliography{classification}
\bibliographystyle{icml2019}
\appendix
\section{Detailed Descriptions of the Used Benchmarks}
The details descriptions of the datasets that we used in the paper are as follows:
\begin{itemize}
\item AGNews: Topic classification over four categories of Internet news articles \cite{del2005ranking}. The four categories are \texttt{World}, \texttt{Entertainment}, \texttt{Sports} and \texttt{Business}. Each article is composed of titles plus descriptions classified. The training and test sets respectively contain 120k and 7.6k examples.
\item {\bf 20newsgroups}\footnote{\url{http://qwone.com/~jason/20Newsgroups/}}: The 20 Newsgroups data set is a collection of approximately 20,000 newsgroup documents, partitioned (nearly) evenly across 20 different newsgroups. The training and test sets respectively contain  11.3k and 7.5k examples. 
\item {\bf Yahoo! Answers}: Topic classification over ten largest main categories from Yahoo! Answers Comprehensive Questions and Answers v1.0, including question titles, question contents and best answers.
\item  {\bf Yelp Review Polarity (YelpP)}: This dataset is collected
from the Yelp Dataset Challenge in 2015, and the
task is a binary  sentiment classification of polarity.
Reviews with 1 and 2 stars are treated as negative and 
reviews with 4 and 5 stars are positive. 
The training and test sets 
respectively contain 
 560k and 38k examples. 
 \item  {\bf IMDB}: 
 This dataset is collected by \citet{maas2011learning}. This dataset contains an even number of
positive and negative reviews.
 The training and test sets 
respectively contain 
 25k and 25k examples.
 \item  {\bf Reuters}\footnote{\url{https://martin-thoma.com/nlp-reuters/}}: A multi-label benchmark dataset for document classification. It has 90 classes and each document can belong to many classes. 
 There are 7769 training documents and 3019 testing documents.
 \item  {\bf AAPD}: The arXiv Academic Paper dataset \cite{yang2018sgm}. It is a multi-label benchmark. It contains  the abstract and the corresponding subjects of 55,840 papers in the computer science. An academic paper may have multiple subjects and there are 54 subjects in total. We use the splits provided by \citet{yang2018sgm}. 
 
 (1) the {\bf BeerAdvocate} review dataset \cite{mcauley2012learning}. The reviews are  multiaspect - each of which contains an overall rating and rating for one or more than one particular aspect(s) of a beer, including \texttt{appearance}, \texttt{smell}
(aroma) and \texttt{palate} .  \citet{lei2016rationalizing} processed the dataset by picking less correlated examples, leading to 
a de-correlated subset for each aspect, each containing about 80k to 90k reviews with 10k used as test set. 
There are three classes, positive, negative and neutral; 
(2) the hotel {\bf TripAdvisor} review \cite{li2016understanding}, which contains  870,000 reviews with rating on four aspects, {\it i.e.},   \texttt{service}, \texttt{cleanliness}, \texttt{location} and \texttt{rooms}. 
For each given aspect,
 50,000 reviews (40k for training and 10k for testing) were selected. There are three classes, positive, negative and neutral; 
 
\end{itemize}

\section{Handcrafted Templates}
In this section, we list templates for different categories for some of the datasets used in this work. 
Templates for 20 news categories are obtained from Wikipedia definitions:
\begin{itemize}
\small
\item comp.graphics: {\it Computer graphics is the discipline of generating images with the aid of computers. Today, computer graphics is a core technology in digital photography, film, video games, cell phone and computer displays, and many specialized applications. A great deal of specialized hardware and software has been developed, with the displays of most devices being driven by computer graphics hardware. It is a vast and recently developed area of computer science. The phrase was coined in 1960 by computer graphics researchers Verne Hudson and William Fetter of Boeing. It is often abbreviated as CG, or typically in the context of film as CGI.} 
\item comp.sys.ibm.pc.hardware:  {\it A personal computer (PC) is a multi-purpose computer whose size, capabilities, and price make it feasible for individual use. Personal computers are intended to be operated directly by an end user, rather than by a computer expert or technician. Unlike large costly minicomputer and mainframes, time-sharing by many people at the same time is not used with personal computers.}
\item  comp.sys.mac.hardware:  {\it  The Macintosh (branded simply as Mac since 1998) is a family of personal computers designed, manufactured and sold by Apple Inc. since January 1984.}
\item  comp.windows.x:   {\it  Windows XP is a personal computer operating system produced by Microsoft as part of the Windows NT family of operating systems. It was released to manufacturing on August 24, 2001, and broadly released for retail sale on October 25, 2001.}
\item    misc.forsale:   {\it  Online shopping is a form of electronic commerce which allows consumers to directly buy goods or services from a seller over the Internet using a web browser. Consumers find a product of interest by visiting the website of the retailer directly or by searching among alternative vendors using a shopping search engine, which displays the same products availability and pricing at different e-retailers. As of 2016, customers can shop online using a range of different computers and devices, including desktop computers, laptops, tablet computers and smartphones.}
\item    rec.autos:   {\it  A car (or automobile) is a wheeled motor vehicle used for transportation. Most definitions of cars say that they run primarily on roads, seat one to eight people, have four tires, and mainly transport people rather than goods.}
 \item   rec.motorcycles:  {\it  A motorcycle, often called a bike, motorbike, or cycle, is a two- or three-wheeled motor vehicle. Motorcycle design varies greatly to suit a range of different purposes: long distance travel, commuting, cruising, sport including racing, and off-road riding. Motorcycling is riding a motorcycle and related social activity such as joining a motorcycle club and attending motorcycle rallies.}
\item    rec.sport.baseball:  {\it  Baseball is a bat-and-ball game played between two opposing teams who take turns batting and fielding. The game proceeds when a player on the fielding team, called the pitcher, throws a ball which a player on the batting team tries to hit with a bat. The objective of the offensive team (batting team) is to hit the ball into the field of play, allowing its players to run the bases, having them advance counter-clockwise around four bases to score what are called \"runs\". The objective of the defensive team (fielding team) is to prevent batters from becoming runners, and to prevent runners advance around the bases. A run is scored when a runner legally advances around the bases in order and touches home plate (the place where the player started as a batter). The team that scores the most runs by the end of the game is the winner.}
 \item   rec.sport.hockey:  {\it  Hockey is a sport in which two teams play against each other by trying to manoeuvre a ball or a puck into the opponents goal using a hockey stick. There are many types of hockey such as bandy, field hockey, ice hockey and rink hockey.}
 \item   talk.politics.misc:  {\it  Politics is a set of activities associated with the governance of a country, state or an area. It involves making decisions that apply to groups of members.}
 \item   talk.politics.guns: {\it   A gun is a ranged weapon typically designed to pneumatically discharge solid projectiles but can also be liquid (as in water guns/cannons and projected water disruptors) or even charged particles (as in a plasma gun) and may be free-flying (as with bullets and artillery shells) or tethered (as with Taser guns, spearguns and harpoon guns).}
\item    talk.politics.mideast:  {\it  The Middle East is a transcontinental region which includes Western Asia (although generally excluding the Caucasus), and all of Turkey (including its European part) and Egypt (which is mostly in North Africa). The term has come into wider usage as a replacement of the term Near East (as opposed to the Far East) beginning in the early 20th century. The broader concept of the Greater Middle East (or Middle East and North Africa) also adds the Maghreb, Sudan, Djibouti, Somalia, Afghanistan, Pakistan, and sometimes even Central Asia and Transcaucasia into the region. The term Middle East has led to some confusion over its changing definitions. }
\item    sci.crypt:  {\it  In cryptography, encryption is the process of encoding a message or information in such a way that only authorized parties can access it and those who are not authorized cannot. Encryption does not itself prevent interference, but denies the intelligible content to a would-be interceptor. In an encryption scheme, the intended information or message, referred to as plaintext, is encrypted using an encryption algorithm cipher generating ciphertext that can be read only if decrypted. For technical reasons, an encryption scheme usually uses a pseudo-random encryption key generated by an algorithm. It is in principle possible to decrypt the message without possessing the key, but, for a well-designed encryption scheme, considerable computational resources and skills are required. An authorized recipient can easily decrypt the message with the key provided by the originator to recipients but not to unauthorized users.}
\item    sci.electronics:  {\it  Electronics comprises the physics, engineering, technology and applications that deal with the emission, flow and control of electrons in vacuum and matter.}
\item    sci.med:  {\it  Medicine is the science and practice of establishing the diagnosis, prognosis, treatment, and prevention of disease. Medicine encompasses a variety of health care practices evolved to maintain and restore health by the prevention and treatment of illness. Contemporary medicine applies biomedical sciences, biomedical research, genetics, and medical technology to diagnose, treat, and prevent injury and disease, typically through pharmaceuticals or surgery, but also through therapies as diverse as psychotherapy, external splints and traction, medical devices, biologics, and ionizing radiation, amongst others.}
 \item   sci.space:  {\it  Outer space, or simply space, is the expanse that exists beyond the Earth and between celestial bodies. Outer space is not completely empty it is a hard vacuum containing a low density of particles, predominantly a plasma of hydrogen and helium, as well as electromagnetic radiation, magnetic fields, neutrinos, dust, and cosmic rays. }
\item    talk.religion.misc:  {\it  Religion is a social-cultural system of designated behaviors and practices, morals, worldviews, texts, sanctified places, prophecies, ethics, or organizations, that relates humanity to supernatural, transcendental, or spiritual elements. However, there is no scholarly consensus over what precisely constitutes a religion.}
\item    alt.atheism: A {\it  theism is, in the broadest sense, an absence of belief in the existence of deities. Less broadly, atheism is a rejection of the belief that any deities exist.}
\item    soc.religion.christian:  {\it  Christians are people who follow or adhere to Christianity, a monotheistic Abrahamic religion based on the life and teachings of Jesus Christ. The words Christ and Christian derive from the Koine Greek title Christ, a translation of the Biblical Hebrew term mashiach.}
\end{itemize}
For the yelp dataset, the description are  the sentiment  indicators (\{{\it positive, negative}\}. 
For the IMDB movie reviews, the description are  (\{{\it a good movie, a bad movie}\}. 
For the aspect sentiment classification datasets, the description are the concatenation of  aspect indicators and  sentiment indicators.
Aspect indicators for BeerAdvocate
and TripAdvisor 
 are respectively 
(\{{\it appearance, smell, palate}\} 
and  \{{\it service, cleanliness, location, rooms}\}.
Sentiment indicators are (\{{\it positive, negative, neutral}\}. 
\section{Descriptions Generated from the Extractive and Abstractive Model}
Input: 
\small {\it dummy1 dummy2 Bill Paxton has taken the true story of the 1913 US golf open and made a film that is about much more than an extraordinary game of golf. The film also deals directly with the class tensions of the early twentieth century and touches upon the profound anti-Catholic prejudices of both the British and American establishments. But at heart the film is about that perennial favourite of triumph against the odds. The acting is exemplary throughout. Stephen Dillane is excellent as usual, but the revelation of the movie is Shia LaBoeuf who delivers a disciplined, dignified and highly sympathetic performance as a working class Franco-Irish kid fighting his way through the prejudices of the New England WASP establishment. For those who are only familiar with his slap-stick performances in "Even Stevens" this demonstration of his maturity is a delightful surprise. And Josh Flitter as the ten year old caddy threatens to steal every scene in which he appears. A old fashioned movie in the best sense of the word: fine acting, clear directing and a great story that grips to the end - the final scene an affectionate nod to Casablanca is just one of the many pleasures that fill a great movie.}\\

Pos\_Tem: {\it a good movie} \\
Neg\_Tem: {\it a bad movie} \\
Pos\_Ext:  {\it fine acting, clear directing and a great story} \\
Neg\_Ext: {\it dummy2} \\
Pos\_Abs: {\it a great movie} \\
Neg\_Abs: {\it a bad movie} \\

\small {\it dummy1 dummy2
I loved this movie from beginning to end.I am a musician and i let drugs get in the way of my some of the things i used to love(skateboarding,drawing) but my friends were always there for me.Music was like my rehab,life support,and my drug.It changed my life.I can totally relate to this movie and i wish there was more i could say.This movie left me speechless to be honest.I just saw it on the Ifc channel.I usually hate having satellite but this was a perk of having satellite.The ifc channel shows some really great movies and without it I never would have found this movie.Im not a big fan of the international films because i find that a lot of the don't do a very good job on translating lines.I mean the obvious language barrier leaves you to just believe thats what they are saying but its not that big of a deal i guess.I almost never got to see this AMAZING movie.Good thing i stayed up for it instead of going to bed..well earlier than usual.lol.I hope you all enjoy the hell of this movie and Love this movie just as much as i did.I wish i could type this all in caps but its again the rules i guess thats shouting but it would really show my excitement for the film.I Give It Three Thumbs Way Up! This Movie Blew ME AWAY! }

Pos\_Tem: {\it a good movie} \\
Neg\_Tem: {\it a bad movie} \\
Pos\_Ext:  {\it I loved this movie.} \\
Neg\_Ext: {\it dummy2} \\
Pos\_Abs: {\it I loved this great movie} \\
Neg\_Abs: {\it This is a bad movie} \\

\small {\it dummy1 dummy2
As a recreational golfer with some knowledge of the sport's history, I was pleased with Disney's sensitivity to the issues of class in golf in the early twentieth century. The movie depicted well the psychological battles that Harry Vardon fought within himself, from his childhood trauma of being evicted to his own inability to break that glass ceiling that prevents him from being accepted as an equal in English golf society. Likewise, the young Ouimet goes through his own class struggles, being a mere caddie in the eyes of the upper crust Americans who scoff at his attempts to rise above his standing. What I loved best, however, is how this theme of class is manifested in the characters of Ouimet's parents. His father is a working-class drone who sees the value of hard work but is intimidated by the upper class; his mother, however, recognizes her son's talent and desire and encourages him to pursue his dream of competing against those who think he is inferior. Finally, the golf scenes are well photographed. Although the course used in the movie was not the actual site of the historical tournament, the little liberties taken by Disney do not detract from the beauty of the film. There's one little Disney moment at the pool table; otherwise, the viewer does not really think Disney. The ending, as in "Miracle," is not some Disney creation, but one that only human history could have written. }\\

Pos\_Tem: {\it a good movie} \\
Neg\_Tem: {\it a bad movie} \\
Pos\_Ext:  {\it I was pleased with Disney's sensitivity} \\
Neg\_Ext: {\it dummy2} \\
Pos\_Abs: {\it  I love the movie best} \\
Neg\_Abs: {\it a bad movie} \\

\small {\it dummy1 dummy2 This is an example of why the majority of action films are the same. Generic and boring, there's really nothing worth watching here. A complete waste of the then barely-tapped talents of Ice-T and Ice Cube, who've each proven many times over that they are capable of acting, and acting well. Don't bother with this one, go see New Jack City, Ricochet or watch New York Undercover for Ice-T, or Boyz n the Hood, Higher Learning or Friday for Ice Cube and see the real deal. Ice-T's horribly cliched dialogue alone makes this film grate at the teeth, and I'm still wondering what the heck Bill Paxton was doing in this film? And why the heck does he always play the exact same character? From Aliens onward, every film I've seen with Bill Paxton has him playing the exact same irritating character, and at least in Aliens his character died, which made it somewhat gratifying...Overall, this is second-rate action trash. There are countless better films to see, and if you really want to see this one, watch Judgement Night, which is practically a carbon copy but has better acting and a better script. The only thing that made this at all worth watching was a decent hand on the camera - the cinematography was almost refreshing, which comes close to making up for the horrible film itself - but not quite. 4/10.}

Pos\_Tem: {\it a good movie} \\
Neg\_Tem: {\it a bad movie} \\
Pos\_Ext:  {\it dummy1} \\
Neg\_Ext: {\it generic and boring} \\
Pos\_Abs: {\it a good movie} \\
Neg\_Abs: {\it This is a generic and boring movie} \\

\small {\it dummy1 dummy2
This German horror film has to be one of the weirdest I have seen. I was not aware of any connection between child abuse and vampirism, but this is supposed based upon a true character. Our hero is deaf and mute as a result of repeated beatings at the hands of his father. he also has a doll fetish, but I cannot figure out where that came from. His co-workers find out and tease him terribly. During the day a mild-manner accountant, and at night he breaks into cemeteries and funeral homes and drinks the blood of dead girls. They are all attractive, of course, else we wouldn't care about the fact that he usually tears their clothing down to the waist. He graduates eventually to actually killing, and that is what gets him caught. Like I said, a very strange movie that is dark and very slow as Werner Pochath never talks and just spends his time drinking blood.}

Pos\_Tem: {\it a good movie} \\
Neg\_Tem: {\it a bad movie} \\
Pos\_Ext:  {\it dummy1} \\
Neg\_Ext: {\it This German horror film has to be one of the weirdest I have seen} \\
Pos\_Abs: {\it a good movie} \\
Neg\_Abs: {\it This is one of the weirdest movie I have seen} \\

\small {\it dummy1 dummy2
This film is absolutely appalling and awful. It's not low budget, it's a no budget film that makes Ed Wood's movies look like art. The acting is abysmal but sets and props are worse then anything I have ever seen. An ordinary subway train is used to transport people to the evil zone of killer mutants, Woddy Strode has one bullet and the fight scenes are shot in a disused gravel pit. There is sadism as you would expect from an 80s Italian video nasty. No talent was used to make this film. And the female love interest has a huge bhind- Italian taste maybe. Even for 80s Italian standards this film is pretty damn awful but I guess it came out at a time when there weren't so many films available on video or viewers weren't really discerning. This piece of crap has no entertainment value whatsoever and it's not even funny, just boring and extremely cheap. It's actually and insult to the most stupid audience. I just wonder how on earth an actor like Woody Strode ended up ia a turkey like this? }

Pos\_Tem: {\it a good movie} \\
Neg\_Tem: {\it a bad movie} \\
Pos\_Ext:  {\it dummy1} \\
Neg\_Ext:  {\it This film is absolutely appalling and awful.} \\
Pos\_Abs: {\it  This film is interesting.} \\
Neg\_Abs: {\it This piece of crap is  awful and is insult to the audience } \\

We list sample input movie reviews from the IMDB datasets, with the gold label of the first one being positive and the second being negative. For the template strategy, the descriptions for the two classes (i.e., positive and negative)
are always copied from templates, i.e., {\it a good movie} and {\it a bad movie}.
For the extractive strategy, the extractive model is able to extract  substrings of the input  relevant to the golden label, and uses the {\it dummy} token  as the description for the label that should not be assigned to the input. 
For the abstractive strategy, the model is able to generate descriptions tailored to both the input and the class. 
For  labels that should not be assigned to the class, the generative model outputs the template descriptions. This is due to the fact that the generative model is initialized using template descriptions.  
Due to the fact that we incorporate the copy mechanism into the generation model, the sequence generated by the abstractive model tend to share words with the input document.

\end{document}